# Explainable AI for Biodiversity Monitoring and Ecological Image Analysis

Brinnae Bent[1], Holly R. Houliston[2,3], Jiayi Zhou[1], Gunel Aghakishiyeva[1], David W. Johnston[4]

[1] Pratt School of Engineering, Duke University

[2] British Antarctic Survey

[3] Department of Applied Mathematics and Theoretical Physics, University of Cambridge

[4] Duke University Division of Marine Science and Conservation, Nicholas School of the Environment, Duke University Marine Laboratory

## Abstract

Artificial intelligence (AI) is rapidly transforming biodiversity monitoring by enabling automated analysis of ecological imagery collected from camera traps, drones, satellites, underwater platforms, and other sensing systems. These tools can expand the scale and speed of conservation assessments, yet many computer vision models remain difficult to inspect, making it challenging to determine whether predictions are based on ecologically meaningful signals or on spurious correlations, sampling biases, and other artifacts that may undermine conservation decisions.

We argue that explainable artificial intelligence (XAI) should become a standard component of ecological model validation because conservation practitioners increasingly depend on understanding not only whether a model is accurate, but why it is accurate. We provide practical guidance for applying XAI to three common ecological computer vision tasks: image classification, object detection, and image segmentation.

To illustrate how XAI can support ecological model auditing, refinement, and deployment, we present two case studies using aerial imagery: harbor seal detection and cetacean anatomical segmentation. These examples demonstrate how explanation methods can identify biologically meaningful cues, reveal false positives driven by background and shape confounds, uncover edge and occlusion effects, and guide data collection, augmentation, and retraining strategies. More broadly, they show how explainability can help assess whether model reasoning aligns with ecological understanding.

We conclude by identifying key challenges and opportunities, including the need for stronger standards for explanation evaluation, improved support for segmentation and foundation models, open-source tools tailored to ecological workflows, and methods capable of explaining spatial, temporal, and multimodal ecological data. By making model behavior more transparent and scientifically interrogable, XAI can help ensure that AI-supported ecological evidence is more reliable, understandable, and actionable for biodiversity conservation.

## 1. Introduction

Ecological research increasingly relies on image data from camera traps, drones, satellites, fixed observatories, and other remote sensing platforms. These systems generate image datasets at spatial and temporal scales that are difficult to analyze manually, creating strong demands for automated computer vision tools. Deep learning models are frequently used to classify species, detect animals or habitat features, count individuals, segment anatomical or landscape structures, and monitor environmental change across large image archives (Norouzzadeh et al., 2018; Tabak et al., 2019; Cubaynes et al., 2019; Duporge et al., 2020; Gray et al., 2022; Buchelt et al., 2024). These applications can substantially accelerate ecological inference and conservation monitoring, but they also introduce a central challenge:

models that perform well according to standard accuracy metrics may still rely on visual cues that are biologically misleading, context-specific, or difficult for human users to evaluate.

This problem is especially important in ecology because image-based models are often deployed under conditions of uncertainty and distribution shift. Ecological imagery varies with season, weather, illumination, sensor type, altitude, habitat structure, animal posture, species interactions, and geographic location. Training datasets are frequently imbalanced, opportunistically sampled, or collected from a limited number of sites (Schneider et al., 2020; Axford et al., 2024). As a result, models may learn shortcuts that reflect the structure of the dataset rather than the ecological target of interest. For example, a classifier may distinguish bird groups using background habitat rather than animal morphology (Parchami-Araghi et al., 2024), or an object detector may mistake ice patches, shadows, wakes, or anthropogenic objects for animals when these features share visual properties with the target class (Beery et al., 2019). Such failures are not always apparent from aggregate performance metrics, yet they can affect downstream estimates of abundance, distribution, habitat use, or population status.

Explainable artificial intelligence (XAI) offers a set of tools for inspecting how models make predictions and for identifying when model behavior conflicts with ecological expectations. In this paper, we use explainability to refer to methods that help characterize why a trained model produced a particular output or pattern of outputs. This differs from interpretability, which generally refers to models whose internal structure is understandable by design, such as simple rule-based models, sparse linear models, or explicitly constrained concept-based models (Rudin, 2019; Molnar, 2022). Many deep learning models used in ecological computer vision, including convolutional neural networks and transformer-based architectures, are not inherently interpretable. For these models, post-hoc XAI methods can provide evidence about the image regions, examples, concepts, or perturbations that influence model behavior. Explanation methods are useful because they can reveal potential failure modes, guide data curation, support model debugging, and help determine whether predictions are consistent with biological knowledge (Figure 1).

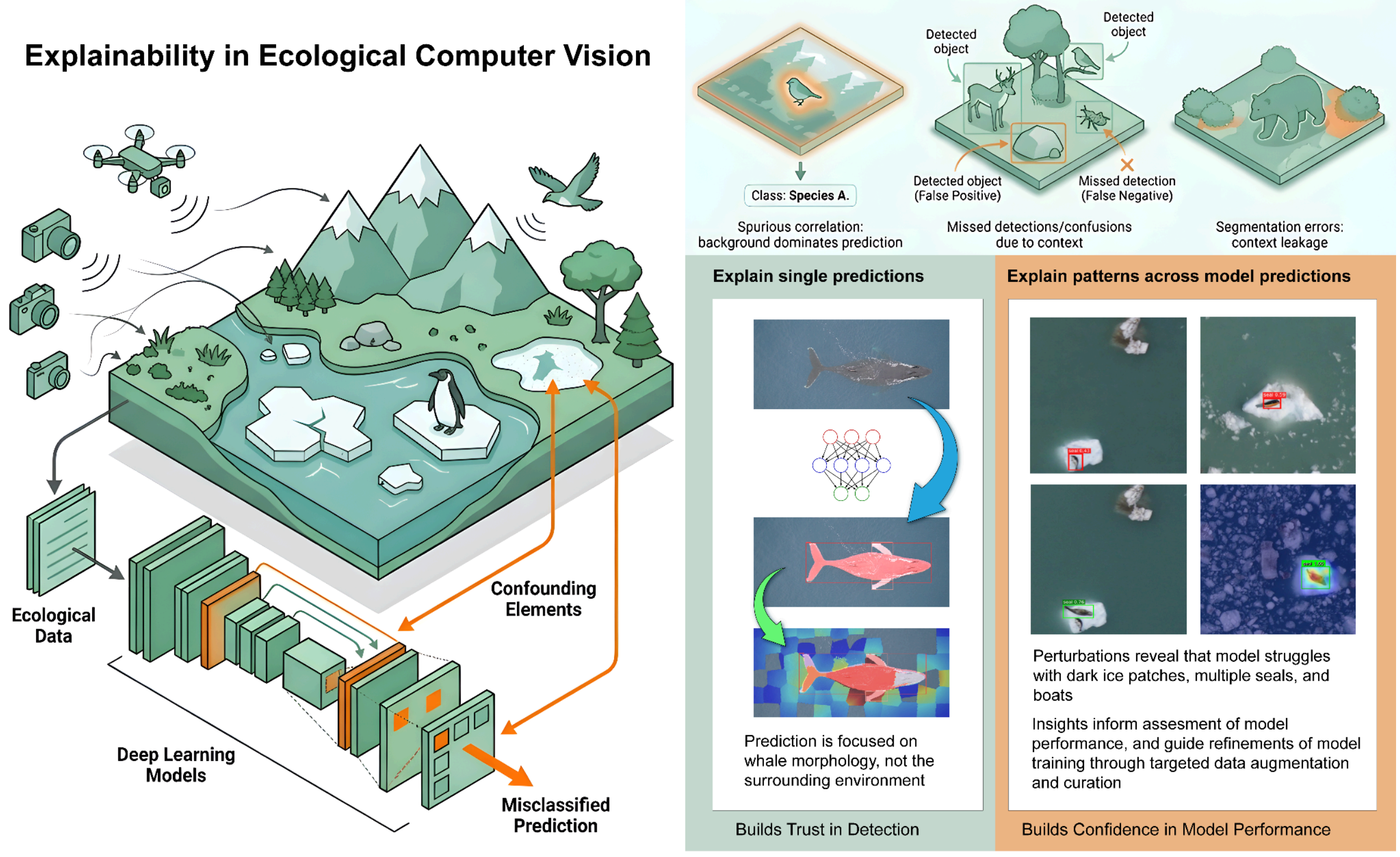

**Figure 1. Graphical Abstract.** Ecological computer vision is applied across challenging settings characterized by diverse species, habitats, sensors, spatial scales, and environmental conditions, where animals may be rare, visually cryptic, partially occluded, or difficult to distinguish from background features. Explainable artificial intelligence (XAI) provides a broad set of approaches for making model behavior more inspectable across ecological applications, including image classification, object detection, and image segmentation. By revealing the evidence, features, context, and data that influence predictions, XAI can help identify biologically meaningful signals, uncover spurious correlations and confounding environmental cues, and diagnose errors such as false positives, missed detections, and segmentation failures. These insights support model auditing, data curation, and iterative model improvement while helping ecologists understand when model predictions are likely to be reliable. As part of a wider framework that includes performance evaluation, uncertainty quantification, robustness testing, and expert review, XAI can promote more transparent and trustworthy use of AI for biodiversity monitoring and conservation.

The artificial intelligence (AI) research community has produced a large body of research on XAI. However, there has been very little crossover between the XAI field and the discipline of ecology. Recent reviews in the earth observation domain highlight emerging AI regulations and guidelines that require algorithms to be explainable and discuss the anticipated impact that XAI will have on the field (Gevaert et al., 2022; Buchelt et al., 2024). Despite recognition of the need for XAI, research on the application of XAI to ecological problems remains scarce, and tools for ecologists and environmental scientists in general are lacking. Most current ecological machine learning (ML) implementations continue to use “black-box” approaches with limited or no explainability, leaving significant gaps in understanding how models make decisions when analyzing ecological imagery for conservation purposes.

Here, we provide practical guidance for applying XAI to three common ecological computer vision tasks: image classification, object detection, and image segmentation. Rather than attempting a comprehensive review of all XAI research, we focus on methods most relevant to ecological image analysis and model auditing. We organize these methods according to common questions ecologists may ask of a model. We then demonstrate these ideas using two aerial-imagery case studies: harbor seal detection and cetacean anatomical segmentation. Finally, we discuss limitations and future directions for ecological XAI.

## 2. Ecology as a High-Stakes, High-Uncertainty Domain

Ecological systems are inherently complex, nonlinear, and context-dependent, with interactions among environmental factors and species operating across varied spatial and temporal scales (Levin, 1998; Dormann et al., 2012). These characteristics make them difficult to model accurately, particularly when using black-box AI tools, whose internal reasoning may not be transparent (Olden et al., 2008; Reichstein et al., 2019). These challenges have real-world implications. For instance, Beery et al. (2019) demonstrated that AI models trained to classify wildlife in camera trap images suffered significant drops in accuracy when applied to new geographic regions, often due to spurious correlations with background features rather than the animals themselves. Likewise, Zurell et al. (2020) reviewed species distribution models and emphasized that uncritical use of climate predictors without ecological validation can lead to misleading outputs. Such risks are heightened when training data are biased or incomplete, a common issue in presence-only datasets (Phillips et al., 2009). Explainability methods enable ecologists to probe model logic and uncover ecologically implausible drivers (Lucas 2020).

The need for XAI is especially pressing in conservation contexts, where AI models increasingly guide policy decisions, resource allocation, and stakeholder engagement. In these scenarios, it is not enough for models to make accurate predictions; decision-makers must also understand the rationale behind those predictions. A common ecological assumption is that globally important environmental variables are also locally influential across different spatial contexts. In other words, if a variable ranks highly in global importance, it is expected to consistently shape predictions across regions. However, Ryo et al. (2021) challenge this assumption by demonstrating that black-box models may rely on different predictors depending on the location. Fernandes et al. (2020) further illustrate this point; their use of XAI techniques revealed that, while climatic variables like temperature seasonality were important in predicting plant species richness, anthropogenic land-use factors – such as proximity to roads and croplands – were often more influential for certain plant families. Furthermore, the relative importance of predictors varied significantly among taxa within the same region, challenging the assumption that species exposed to similar environmental conditions respond uniformly. Together, these studies demonstrate how XAI enhances ecological modeling by revealing hidden model behaviors, uncovering bias, and refining hypotheses for future research. Beyond improving technical validity, explainability also fosters transparency, reproducibility, and public trust, particularly in high-stakes conservation decisions. In domains characterized by uncertainty and ecological risk, XAI is essential to ensure that ML complements, rather than replaces, sound ecological reasoning (Fernandes et al., 2020; Ryo et al., 2021).

## 3. Conceptual Taxonomy for Explainable Ecological Computer Vision

Explainability methods differ in what they explain, how they generate explanations, and what kinds of claims they can support. Here we present a taxonomy for explainability in ecological computer vision.

A first distinction is between interpretable-by-design models and post-hoc explanations. Interpretable-by-design models have structures that are intended to be understandable without additional explanation tools, such as sparse linear models, decision trees, generalized additive models, prototype networks, or concept-based models (Molnar, 2022). These approaches can be valuable when ecological decisions require transparent reasoning from the outset (Rudin, 2019). However, many ecological image-analysis workflows use complex deep learning models. For these models, explanations are typically generated post hoc, after the model has already been trained. Post-hoc methods do not make the model inherently interpretable, but they can help users inspect model behavior, identify potential failure modes, and evaluate whether predictions align with ecological expectations (Molnar, 2022).

A second distinction is between local explanations and global or dataset-level explanations. Local explanations describe why a model produced a particular prediction for a particular image, detection, or segmentation mask. For example, a local explanation may show which pixels or image regions contributed to classifying, detecting, or segmenting an animal. These explanations are useful for investigating false positives, false negatives, uncertain predictions, or cases of ecological interest. Global explanations, by contrast, summarize patterns across many predictions. In ecological applications, this may involve aggregating attribution maps across a test set, comparing explanations across habitats or seasons, measuring how often attribution falls inside target objects rather than background regions, or identifying recurring confounds such as shadows, snow, wakes, vegetation, or sensor artifacts. Because ecological models are often used to support inference across space and time, dataset-level explanations are essential complements to single-image examples.

A third distinction is between model-specific and model-agnostic methods. Model-specific methods require access to internal model components such as gradients, feature maps, attention weights, or learned representations. Class activation mapping methods, including Grad-CAM (Selvaraju et al., 2020), Grad-CAM++ (Chattopadhay et al., 2018), LayerCAM (Jiang et al., 2021), HiResCAM (Draelos et al., 2020), and EigenGradCAM (Muhammad et al., 2020), are examples of model-specific approaches commonly applied to convolutional neural networks. These methods can be computationally efficient and visually informative, but they depend on architectural details and may require modification for object detection, segmentation, or transformer-based models. Model-agnostic methods rely only on model inputs and outputs, without requiring access to the internal architecture or parameters. LIME (Ribeiro et al., 2016), Anchors (Ribeiro et al., 2018), image perturbation tests (Fong et al., 2017), and some counterfactual approaches (Goyal et al., 2019) fall into this category. These methods can be applied more broadly, but they may be computationally expensive and sensitive to choices about how images are masked, segmented, or modified.

Explanation methods can also be grouped by the type of evidence they provide. Attribution methods estimate which pixels, regions, or features most influenced a prediction. These include saliency maps, Integrated Gradients, and CAM-based approaches such as Grad-CAM, LayerCAM, and HiResCAM (Selvaraju et al., 2020; Jiang et al., 2021; Draelos et al., 2020;

Sundararajan et al., 2017; Simonyan, 2014). Perturbation methods alter or remove parts of an image and measure how predictions change, helping test whether a model depends on an organism, background feature, or contextual cue (Fong et al., 2017; Aghakishiyeva et al., 2025). Example-based methods explain predictions by retrieving similar training or reference examples, which can support expert review of rare species, ambiguous detections, or unusual segmentation outputs (Kim et al., 2016). Concept-based methods evaluate whether human-defined concepts, such as flipper shape, body curvature, canopy texture, or water disturbance, are associated with model predictions. Counterfactual explanations ask how an image would need to change for the model to produce a different output, such as replacing a seal-like ice patch or removing an animal from the scene (Goyal et al., 2019). Influence-based methods trace predictions back to training examples that most affected the model, which can help identify mislabeled data, class imbalance, or site-specific bias (Koh, 2017). Attention-based explanations use attention weights or derived attention-attribution methods to inspect how transformer-based or attention-based models distribute focus across image regions or tokens. Attention-based methods are increasingly relevant, but raw attention should be interpreted cautiously because attention weights do not necessarily provide faithful causal explanations of model behavior (Bastings and Filippova, 2020; Jain and Wallace, 2019; Wiegreffe and Pinter, 2019).

Finally, explanations should be evaluated in terms of both plausibility and faithfulness. Plausibility refers to whether an explanation appears reasonable to a human observer or domain expert. For example, an attribution map that highlights the body outline of a whale may be plausible because it aligns with ecological intuition. Faithfulness refers to whether the explanation accurately reflects the information the model actually used to produce its output. A plausible explanation is not necessarily faithful: a heatmap may look biologically meaningful while failing to capture the true decision process of the model. Conversely, a faithful explanation may reveal unexpected or unintuitive behavior, such as reliance on water texture, shadow, or image borders. For this reason, ecological applications of XAI should not rely on visual inspection alone. Explanations should be tested through perturbations, sanity checks, comparisons across error types, and expert evaluation whenever possible.

In the sections that follow, we use this framework to compare explanation methods for three common ecological computer vision tasks: image classification, object detection, and image segmentation.

## 4. Explainability Recommendations for Ecological Applications

Below, we provide recommendations for applying XAI methods to three computer vision tasks used frequently in ecological research: image classification, object detection, and image segmentation. Table 1 lists current explainability methods and the computer vision task(s) for which the method is appropriate.

**Table 1:** Appropriate application of artificial intelligence explainability methods for different computer vision tasks (image classification, object detection, image segmentation).

| Explainability Method | Image Classification | Object Detection | Image Segmentation |
|---|---|---|---|
| GradCAM (Selvaraju, 2020) | x | x | x |
| GradCAM++ (Chattopadhay, 2018) | x | x | |
| LayerCAM (Jiang, 2021) | x | x | x |
| HiResCAM (Draelos, 2020) | x | x | x |
| EigenGradCAM (Muhammad, 2020) | x | x | |
| Saliency Maps (Simonyan, 2014) | x | | |
| Integrated Gradients (Sundararajan, 2017) | x | x | x |
| LIME (LIME variants/Anchors) (Ribeiro, 2016; 2018) | x | x | x |
| Prototype-based Explanations (Kim, 2016) | x | x | |
| Image Perturbations (Fong, 2017) | x | x | x |
| Counterfactuals (Goyal, 2019) | x | x | |
| Influential Instances (Koh, 2017) | x | | |
| Feature Visualization (Olah, 2017) | x | x | x |
| Network Dissection (Bau, 2017) | x | | |
| Concept Activation Vectors (TCAV) (Kim, 2018) | x | | |

## 4.1 Common Explanation Methods

Before selecting an explainability method for a specific ecological computer vision task, it is useful to understand the basic logic behind the major explanation methods.

Class activation mapping methods, including Grad-CAM (Selvaraju et al., 2020), Grad-CAM++ (Chattopadhay et al., 2018), LayerCAM (Jiang et al., 2021), HiResCAM (Draelos et al., 2020), and EigenGradCAM (Muhammad et al., 2020), use internal feature maps from convolutional layers to identify image regions associated with a model output. In general, these methods combine feature activations with gradient or activation-based weights to produce a heatmap showing where the model focused when making a prediction. CAM methods are especially useful when the question is spatial: for example, whether a species classification, bounding box, or segmentation output is driven by the organism itself or by background features such as vegetation, water texture, ice, shadows, or sensor artifacts.

LIME (Ribeiro et al., 2016) and Anchors (Ribeiro et al., 2018) are model-agnostic local surrogate methods. Rather than inspecting model internals, these approaches perturb parts of an input image and observe how the model output changes. LIME then fits a simpler, inherently interpretable model around the prediction to estimate which regions contributed most strongly to the output. Anchors identify if-then rules for local explanations. In ecological imagery, these methods can be useful for investigating uncertain predictions, false positives, false negatives, or cases where researchers want to understand the importance of particular image regions without relying on gradients.

Integrated Gradients is a gradient-based attribution method that estimates feature importance by accumulating gradients along a path from a baseline input to the observed input (input image) (Sundararajan et al., 2017). In images, the baseline might be a black image, a blurred image, or other reference states. The method assigns importance to pixels according to how much they contribute as the input changes from the baseline to the real image. Integrated Gradients can be useful when researchers want a more formal attribution method than simple saliency maps, although results can depend on the choice of baseline.

Perturbation (Fong et al., 2017) and counterfactual methods (Goyal et al., 2019) evaluate model behavior by altering the image and measuring how predictions change. Perturbations may involve masking, blurring, occluding, replacing, or inpainting image regions (Aghakishiyeva et al., 2025). Counterfactuals ask what change would be needed to alter the model's prediction, such as removing an animal, replacing a background feature, or changing a potential confound. These methods are particularly useful in ecology because they can test whether a model depends on biologically meaningful morphology or on contextual shortcuts.

Prototype and example-based methods explain predictions by comparing an input image to similar learned or retrieved examples (Kim et al., 2016). Instead of asking which pixels were important, these methods ask what the prediction resembles. This can be valuable for expert review, especially for rare species, ambiguous detections, or unfamiliar model outputs. Example-based explanations may help researchers identify whether a model is generalizing from appropriate reference cases or relying on visually similar but ecologically irrelevant examples.

Influence-based methods estimate which training examples most affected a model prediction (Koh et al., 2017). These approaches can help trace surprising outputs back to influential training samples, potentially revealing mislabeled images, duplicated data, site-specific bias, or underrepresented classes. For ecological datasets, where training data are often imbalanced or collected opportunistically, influence methods may be especially useful for diagnosing whether errors arise from model architecture, annotation quality, or dataset composition.

Concept-based methods explain models using human-defined concepts rather than individual pixels. Approaches such as Testing with Concept Activation Vectors (TCAV) evaluate whether concepts like body curvature, fin shape, canopy texture, snow cover, water disturbance, or habitat structure are associated with model predictions (Kim et al., 2018). Post-hoc concept bottleneck models (Yuksekgonul et al., 2022) go further by explicitly linking human-interpretable concepts to the model's prediction pathway. These approaches can be especially useful when ecologists want explanations that map directly onto biological traits or environmental variables, although they require careful definition and annotation of relevant concepts.

Attention-based explanations are increasingly relevant as transformer-based models become more common in ecological computer vision. Attention mechanisms assign weights to relationships among image patches, tokens, or regions, and these weights can sometimes help identify what parts of an image influenced a model output (Bastings and Filippova, 2020). However, raw attention should be interpreted cautiously: attention weights are not automatically faithful causal explanations of model behavior (Jain and Wallace, 2019; Wiegreffe and Pinter, 2019). More robust approaches often combine attention methods with gradients, perturbations, rollouts, or other attribution techniques (Bastings and Filippova, 2020). For ecological applications, attention-based explanations are promising but should be validated against perturbation tests, expert review, and task-specific failure analyses.

Together, these method families provide complementary forms of evidence. Attribution and CAM methods are useful for visual localization, perturbation and counterfactual methods for testing sensitivity, example-based methods for comparison with known cases, influence methods for tracing training-data effects, concept methods for linking predictions to ecological traits, and attention-based methods for inspecting newer architectures. In practice, ecological model auditing will often require combining several explanation types rather than relying on a single method.

### 4.2 Classification

Classification tasks in ecology involve assigning a label to an entire image; for example, identifying a species in a photograph, categorizing habitat types, or predicting land cover classes. Deep learning models have achieved high accuracy in these contexts, such as classifying animal species in large camera trap datasets (Norouzzadeh et al., 2018; Tabak et al., 2019) or recognizing species in smaller ecological surveys (Islam et al., 2023). Image classification has also been applied to environmental imagery beyond wildlife, including mapping land cover and classifying elements of ecological systems across landscapes using satellite imagery (Fan et al., 2023; Sunde et al., 2024). The utility of these models depends heavily on understanding why a particular sample was assigned to a given class, since classification often relies on subtle features such as color patterns, shape, texture, or seasonal cues. Rare or threatened species pose particular challenges because they are typically underrepresented in datasets, creating a class imbalance that can degrade performance (Norman et al., 2023). In such cases, it is critical to confirm that predictions are based on biologically meaningful signals rather than spurious correlations.

A well-known challenge in image classification is achieving robust performance under environmental conditions that are underrepresented or absent from the training data. In these situations, models may rely on background features that are correlated with species occurrence but do not directly reflect the broader range of features that govern the presence of the species itself. Such spurious associations can reduce the ability of CNNs to generalize across habitats, regions, and observation conditions (Norman et al., 2023). Improving model explainability helps ensure that predictions are driven by ecologically meaningful characteristics rather than isolated background patterns, increasing confidence in automated classifications and their reliability for ecological applications (Islam et al., 2023; Sunde et al., 2024).

**Table 2:** Explainability in Ecological Image Classification

| Monitoring Need | Question | Recommended Explainability Method | When to Use | Why This Works | Ecology-Specific Uses and Considerations |
|---|---|---|---|---|---|
| Verify why a species was classified | "Why did the model classify this as species X?" | GradCAM | Best for large, centered subjects with clear features | Highlights image regions driving the classification | Use to quickly verify camera trap or drone image predictions |
| | | GradCAM++ | Better for small, overlapping, or co-located species | Handles complex scenes and fine distinctions | Especially helpful for animals in dense scenes |
| | | LayerCAM | When species differences are subtle (e.g., feather edges, ear tufts) | Captures fine-grained detail | Useful for distinguishing visually similar species |
| | | HighResCAM | High-resolution images or small species | Produces precise and localized attribution | Best for aerial imagery or detailed species traits |
| Understand visual focus | "What features drive classification?" | EigenGradCAM | When prediction is uncertain or multiple | Visualization not tied to specific class | Use when model is confused between species or habitat types |

| | | | species are plausible | | |
|---|---|---|---|---|---|
| | | Saliency Maps (e.g., vanilla, SmoothGrad) | When seeking basic pixel-level sensitivity | Highlights input regions that most affect output | Often noisy; good for quick inspection or comparison |
| Find similar examples | “What does this resemble?” | Prototype Matching | When validating rare detections or expert review | Surfaces nearest neighbor images from training or known reference sets | Builds trust in species detection by showing comparable examples |
| Investigate model uncertainty | “Why is this borderline or misclassified?” | Local Interpretable Model-agnostic Explanations (LIME) | For borderline predictions or complex backgrounds | Measures importance of image regions via perturbation | Useful for understanding false positives |
| Understand classification patterns across data | “What patterns does the model rely on?” | Aggregate CAM heatmaps | For auditing model focus on trends across a dataset | Reveals consistent attention to features or confounding background | Identify dataset bias (e.g., species associated with vegetation type) |
| | | Feature Visualization | When exploring what specific neurons have learned | Generates synthetic images that activate filters | Helps interpret whether filters correspond to species features |
| | | | | | (e.g., stripes, claws) |
| | | Network Dissection | For interpreting entire networks conceptually | Labels filters with known visual concepts (e.g., “fur”, “sky”) | Requires a concept-labeled dataset, which is difficult to curate |
| Test robustness | “When does classification fail?” | Image Perturbation / Counterfactuals | Stress test model under varying conditions | Reveals sensitivity to occlusion, lighting, and habitat | Helps ensure generalization across different environments or camouflage |
| Trace influence of training data | “What training data caused this prediction?” | Influential Instances | When debugging surprising or incorrect classifications | Links prediction to the most influential training samples | Can help spot mislabeled or imbalanced training data |

### 4.3 Object Detection

Object detection is a fundamental task in computer vision that involves identifying and localizing specific objects within an image by drawing bounding boxes around them and assigning class labels (e.g., “seabird,” “whale”). Unlike image classification, which assigns a label to an entire image, object detection provides spatial information about where each object is located within the image. In ecological research, object detection is widely used to automate tasks such as identifying individual animals in camera trap footage (Norouzzadeh et al., 2018), counting trees or other vegetation in aerial or satellite imagery (Brandt et al., 2020), and monitoring invasive species or habitat changes over time (Lake et al., 2022). These applications help scale biodiversity assessments, improve the efficiency of fieldwork, and provide timely data for conservation planning (Tabak et al., 2019). However, for object detection to be trusted and actionable in ecological contexts, it must be explainable. Researchers need to understand why the model labeled a region as a specific object and how it made localization decisions. This includes analyzing which visual features (e.g., color, texture, shape) influenced the model's predictions. Explainability is particularly important in low-sample settings or for detecting rare, behaviorally cryptic, or camouflaged species, where false positives and false negatives can have disproportionate ecological impacts.

**Table 3:** Explainability in Ecological Object Detection

| **Monitoring Need** | **Question** | **Recommended Explainability Method** | **When to use** | **Why this works** | **Ecology-specific uses and considerations** |
|---|---|---|---|---|---|
| Verify why a species was detected | "Why did the model detect this species?" | GradCAM | Best for large, clear objects with strong class signals | Basic heatmap from classification head; quick sanity check | Use for broad overview or QA checks |
| | | GradCAM++ | Better for overlapping or small objects | Handles multiple object instances better | Useful for groups or clusters of animals |
| | | LayerCAM | Best when detail matters (e.g., | Preserves fine-grained | Good for species with subtle visual |
|  |  |  | beak vs. tail) | localization at multiple layers | cues |
|  |  | HighResCAM | For high-resolution drone/satellite imagery | Produces sharper maps with better spatial alignment | Helps when species are small in frame or partially occluded |
| Understand visual focus | “What features drive detection?” | EigenGradCAM | When you want a class-independent explanation (general model focus) | Decomposes signal from all classes | Use when you don’t trust the class label but still want to see model focus |
| Find similar examples | “What does this resemble?” | Prototype Matching | When doing expert review or spotting errors | Builds trust through comparison to known cases | Especially good for rare species detections |
| Investigate model uncertainty | “Why is this borderline?” | Local Interpretable Model-agnostic Explanations (LIME) and LIME variants | For debugging false positives/negatives | Measures effect of masking parts of the image on confidence | Helps in reviewing model decisions in ambiguous scenes |
| Understand detection patterns across data | “What patterns does the model pick up across images?” | Aggregate CAM heatmaps | Best for model auditing | Useful for quality assessment and model improvement | Look at trends across species, terrain types, or weather |
| Test robustness | “When does detection fail?” | Image Perturbation / Counterfactuals | When studying model brittleness | Helps discover sensitivity to background, lighting | Inform labeling or post-processing policies |

### 4.4 Segmentation and Measurement Tasks

Segmentation and measurement tasks involve dividing an image into meaningful regions to extract spatial and quantitative ecological information, such as delineating individual plant or tree boundaries (Weinstein et al., 2019), measuring canopy cover (Zhou et al. 2025), identifying intertidal habitats (Ridge et al 2020), assessing landform changes (Gray et al. 2021), or mapping coral health using automated annotation of reef imagery (González-Rivero et al., 2020). These tasks are important in remote sensing and field-based image analysis, where accurate boundaries and area-based measurements directly inform ecological metrics like biomass, species distribution, or habitat quality. To make these models explainable, it is essential to understand *why* a region was segmented in a particular way and *what* visual or spectral features influenced the boundary decisions or measurements. Unlike object detection, which identifies discrete entities, segmentation requires spatial coherence: explanations must go beyond isolated pixel importance and account for regional continuity that matches natural structures. Errors in segmentation can propagate into downstream analyses, leading to over- or underestimation of canopy cover, misclassification of coral health zones, and, ultimately, flawed ecological conclusions. This is particularly important for fine-scale segmentation tasks, where even small spatial errors can substantially affect growth estimates. Explanations must also address disagreements between model output and expert annotations, especially in edge cases where boundaries are ambiguous due to shadow, occlusion, or gradual transitions in natural features.

**Table 4:** Explainability in Ecological Image Segmentation

| **Monitoring Need** | **Question** | **Recommended Explainability Method** | **When to Use** | **Why This Works** | **Ecology-Specific Uses and Considerations** |
|---|---|---|---|---|---|
| Verify why a region was segmented | “Why did the model segment this as class X?” | LayerCAM | Best when evaluating fine-scale boundaries or morphology | Captures detailed, spatially aware activations across layers | Useful for segmenting nests, coral, vegetation edges, or shoreline boundaries |
| | | HighResCAM | When working with high-resolution aerial or satellite imagery | Produces sharper, more spatially accurate maps | Best for small features or species in high-resolution data |
| Understand model attention | “What parts of the image influence segmentation?” | Aggregate CAM heatmaps (LayerCAM/HighResCAM) | When auditing a model’s attention across many samples | Visualizes systematic focus regions across a dataset | Helps identify over-reliance on background features like shadows or terrain |
| Investigate model uncertainty | "Why is this segmentation questionable?" | LIME (with patch masking over the input) | When exploring how small changes affect predicted mask | Highlights image patches that change the segmentation outcome | Good for diagnosing errors around class boundaries or in mixed-habitat images |
| Test robustness | "When does segmentation fail?" | Image Perturbation / Counterfactuals | To test stability under occlusion, lighting, or context shifts | Simulates real-world variability (e.g., cloud cover, vegetation growth) | Important for assessing model reliability across seasons, environments, or resolutions |

### 4.5 From Prediction-Level to Dataset-Level Explanations

Many applications of XAI focus on individual predictions: why a particular image was classified as one species, why a bounding box was drawn around a suspected animal, or why a segmentation mask included a particular region. These prediction-level explanations are valuable because they allow researchers to inspect specific successes, failures, and uncertain outputs (Ribeiro et al., 2016; Selvaraju et al., 2020; Molnar, 2022). They are especially useful for understanding false positives, false negatives, rare detections, edge cases, and examples that are ecologically important.

However, ecological models are rarely used to interpret a single image in isolation. They are usually deployed across many images, sites, seasons, species, or environmental conditions. For this reason, prediction-level explanations should be complemented with dataset-level explanation analyses that summarize model behavior across groups of samples (Nauta et al., 2023). Dataset-level explanations can reveal recurring patterns that may not be visible from individual examples, including systematic reliance on background habitat, image borders, lighting conditions, water texture, snow or ice patterns, shadows, vegetation structure, or sensor artifacts. These patterns are important because deep learning models can learn shortcuts that perform well within a dataset but fail under distribution shift (Beery et al., 2018; Geirhos et al., 2020).

One practical approach is to aggregate attribution maps across many predictions. For example, CAM-based heatmaps can be averaged or summarized across true positives, false positives, false negatives, species groups, sites, or habitat types (Selvaraju et al., 2020; Nauta et al., 2023). Aggregated explanations can reveal whether a model consistently focuses on the target organism, on the surrounding habitat, or on recurring non-target features. In object detection and segmentation workflows, attribution maps can also be compared with annotated bounding boxes or masks. Measuring the proportion of attribution that falls inside versus outside target boxes or masks provides a simple diagnostic for whether the model is primarily using organismal features or contextual information. Related localization-based evaluation strategies are commonly used to assess whether saliency maps align with relevant image regions (Samek et al., 2017; Petsiuk et al., 2021). This analysis is particularly useful when models appear accurate but may be relying on shortcuts that could fail under distribution shift.

Dataset-level explanation analyses can also be stratified by ecological and technical covariates. For aerial imagery, explanations may be compared across altitude, image resolution, glare, sea state, substrate type, ice or water texture, lighting, animal density, occlusion, and sensor platform. For camera-trap imagery, analyses might be stratified by habitat, time of day, season, camera location, background vegetation, snow cover, or species identity. For segmentation tasks, explanations can be compared across boundary types, object sizes, partial occlusion, edge crops, and mixed-habitat scenes. Stratifying explanations in this way can help determine whether a model behaves consistently across ecological conditions or whether it depends on context-specific visual cues (Beery et al., 2018; Geirhos et al., 2020).

Perturbation results can also be summarized across a test set rather than shown only as individual examples. For instance, researchers can measure how often predictions change after masking the animal, replacing the background, altering brightness, adding blur, removing water disturbance, or inserting hard-negative objects. These summaries provide a more systematic measure of model sensitivity and can help identify which environmental or visual changes most affect model outputs (Fong et al., 2017; Samek et al., 2017). When paired with conventional performance metrics, perturbation summaries can distinguish models that are accurate for robust reasons from models that are accurate only under narrow visual conditions.

Finally, dataset-level explanations should include expert ecological review whenever possible. Domain experts can assess whether highlighted regions correspond to biologically meaningful features or whether explanations instead emphasize implausible cues such as shadows, wakes, image edges, or artifacts. Expert review is especially important because visually plausible explanations are not necessarily faithful to the model's decision process, and XAI evaluation should include application-grounded assessment where possible (Doshi-Velez and Kim, 2017; Adebayo et al., 2018; Nauta et al., 2023; Suh et al., 2025).

Together, prediction-level and dataset-level explanations allow XAI to support both local debugging and broader ecological validation. Individual examples help diagnose specific model behaviors, while aggregate analyses reveal whether those behaviors are consistent across the range of conditions in which the model will be used.

## 4.6 Emerging Trends

Recent ecological computer vision workflows increasingly use transformer-based and foundation models, which create new opportunities for ecological monitoring but also complicate explainability (Meier and Lüddecke, 2025; Gálvez-Salido et al., 2026; Kyathanahally et al., 2022). Unlike conventional CNN workflows, these models often rely on attention mechanisms, patch-token representations, text prompts, or prompt-conditioned masks. As a result, explanation methods developed for CNNs do not always transfer directly.

Models such as the Segment Anything Model (SAM) (Kirillov et al., 2023), Grounding DINO (Liu et al., 2023), and CLIP-like vision-language models (Radford et al., 2021) are becoming relevant to ecological image analysis (Gálvez-Salido et al., 2026). These models can support zero-shot or few-shot workflows, open-vocabulary detection, prompt-based segmentation, and image-text retrieval. This is useful in ecological settings where labeled data are scarce or where users need to identify rare, novel, or taxonomically diverse targets. However, their explanations require special care because predictions may depend on prompt wording, text-image alignment, pretraining data, and domain mismatch between general web-scale imagery and ecological imagery (Fantozzi and Naldi, 2024).

For transformer-based models, attention maps can provide useful diagnostic information about relationships among image patches or tokens (Vig, 2019), but raw attention weights should not be treated as definitive explanations. Prior work has shown that attention weights may not reliably identify the causal features responsible for a prediction, and different attention patterns can sometimes produce similar outputs (Jain and Wallace, 2019; Wiegreffe and Pinter, 2019). More informative approaches include attention rollout and attention flow, which estimate how information propagates through multiple transformer layers, and transformer attribution methods that combine attention with gradients or relevance propagation (Abnar and Zuidema, 2020; Chefer et al., 2021). For ecological applications, these methods may help evaluate whether a model attends to organisms, habitat structure, water texture, shadows, or other contextual features, but they should be validated with perturbation tests and expert review.

Concept-based methods also offer a way to make explanations more ecologically meaningful. Rather than explaining predictions only through pixels or heatmaps, methods such as TCAV (Kim et al., 2018) and concept bottleneck models evaluate the role of human-defined concepts in model decisions (Yuksekgonul et al., 2022). In ecological imagery, relevant concepts might include body curvature, flipper shape, antler presence, plumage pattern, canopy texture, water disturbance, snow cover, or habitat structure. These approaches are attractive because they can align model explanations with biological traits or environmental variables, but they require careful concept definition, annotation, and validation to avoid replacing pixel-level opacity with poorly specified ecological concepts.

For prompt-based segmentation models, such as SAM, explanations should also include tests of prompt sensitivity and mask stability. Because these models can produce different masks depending on point prompts, box prompts, mask prompts, or prompt placement (Jiang, 2025), users should evaluate whether ecologically meaningful segmentations remain stable under small prompt changes. In ecological workflows, this may involve testing whether whale body masks, coral boundaries, tree crowns, nests, or shoreline features remain consistent across plausible prompts, image resolutions, and environmental conditions. Instability across prompts can reveal ambiguous boundaries, weak domain transfer, or overreliance on background structure. Therefore, for segmentation foundation models, prompt sensitivity and mask stability should be treated as part of the explainability and validation workflow rather than as purely technical implementation details.

Emerging approaches in mechanistic interpretability, including techniques like dictionary learning with sparse autoencoders and visual steering (Joseph et al., 2025; Ranaldi 2025), aim to dissect the internal representations of transformer models. These methods attempt to uncover how attention heads, token interactions, and intermediate representations contribute to specific predictions, offering the potential for more grounded and scientifically aligned insights. However, incorporating these methods into ecological computer vision workflows remains challenging due to their substantial computational complexity, large memory requirements, and difficulties scaling mechanistic analyses to high-resolution imagery, multimodal sensor streams, and increasingly large foundation models commonly used in ecology.

## 5. Case Studies: Seal Monitoring and Cetacean Anatomical Region Segmentation

To illustrate how explainability may be applied in the context of ecological research, we demonstrate its use in two tasks: pinniped monitoring via object detection and cetacean body part identification via segmentation. Code and data is provided in the Supplemental Material.

### 5.1 Case Study: Seal Monitoring

To illustrate explainability techniques on different object detection models, we trained both a Faster-RCNN object detection model (Ren, et al., 2015) (Appendix A.2.1) and a YOLOv9 object detection model (Wang et al., 2024) (Appendix A.2.2) on a dataset of 1,974 drone images taken at a single site in Glacier Bay National Park (Appendix A.1). To identify attributions (where the model focused when predicting a bounding box), we used gradient-based class activation mapping (Appendix A.4.1), selecting HiResCAM and LayerCAM as the dataset consisted of high resolution drone images and the animals to be detected were relatively small. To investigate model uncertainty (specifically, false positives), we used LIME (Appendix A.4.2), and to examine model robustness, we probed the model with perturbations (Appendix A.4.3).

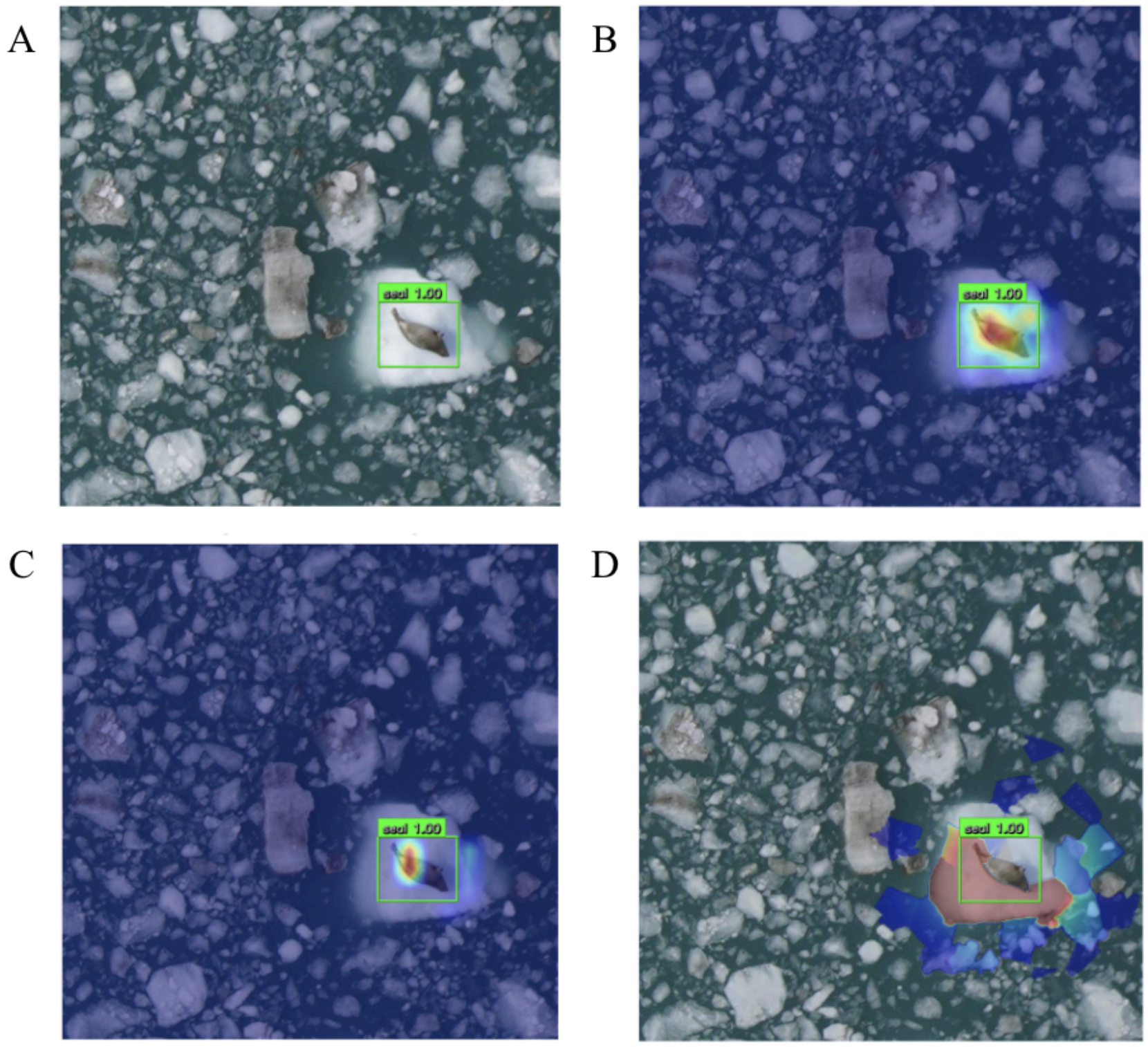

**Figure 2.** (A) Original Faster-RCNN detection of a true positive harbour seal in a drone image. (B) LayerCAM attribution map highlighting image regions supporting the bounding box prediction. (C) HiResCAM attribution with spatial alignment to the object. (D) LIME shows superpixels with the highest estimated contribution to the prediction.

Figure 2 depicts a true positive harbor seal detection using the Faster-RCNN model. Regions of high attribution by the model align with human expectations, indicating the prediction is driven from the animal itself rather than by background features. Both LayerCAM (Figure 2B) and HiResCAM (Figure 2C) highlight that the region of highest attribution is the lower body of the seal, where the curvature of the rear flippers begin. LIME (Figure 2D) corroborates the gradient-based explanations, illustrating that the superpixel containing the seal has the highest estimated contribution to the prediction. Explanations of correct predictions, like Figure 2, can build trust in the model. Pairing traditional evaluation performance metrics with explainability techniques like this can support stakeholders in their decisions to use models in ecological research and conservation management.

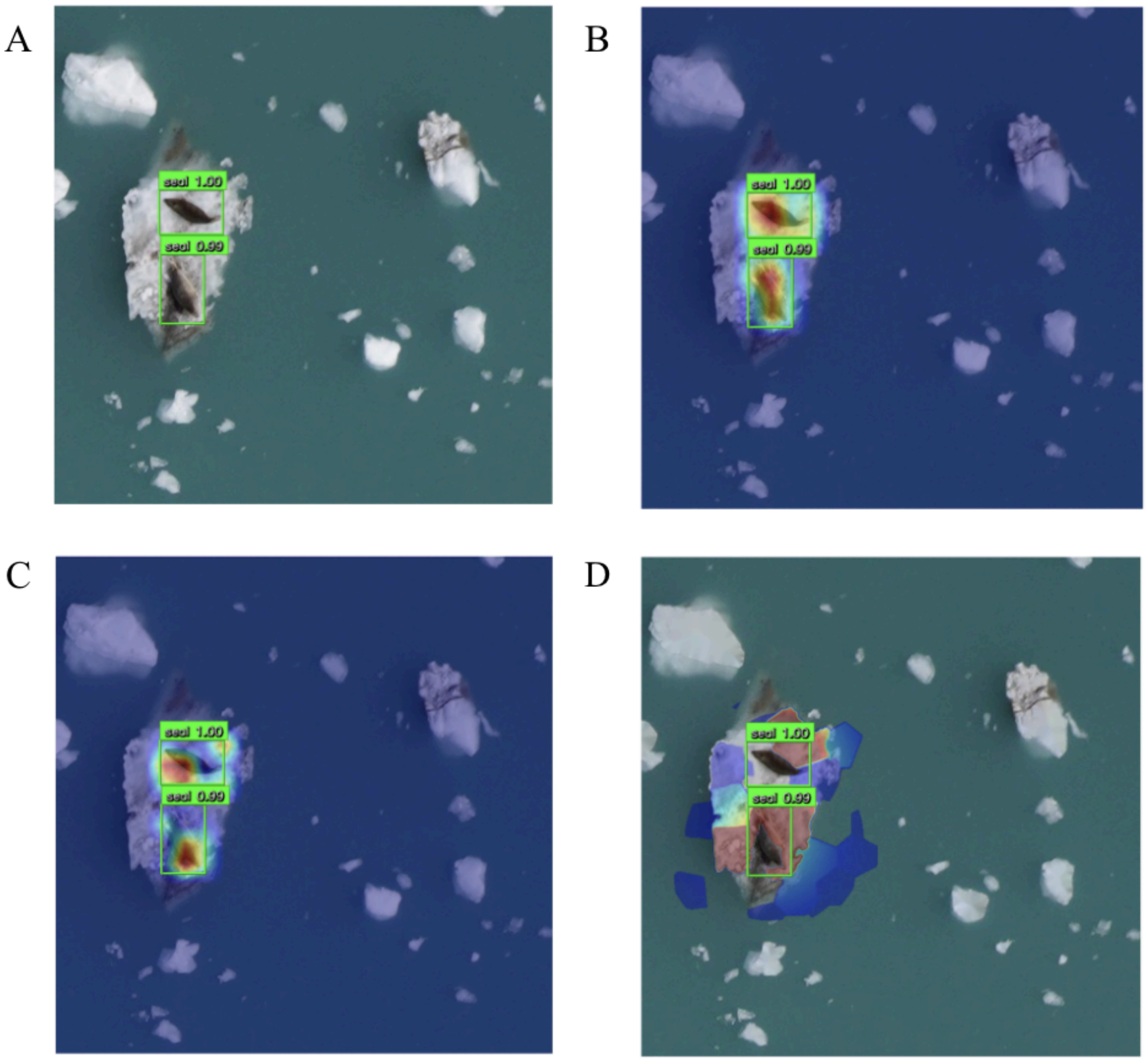

**Figure 3.** (A) Original Faster-RCNN model detection of a true positive and false positive harbour seal in a drone image. (B) LayerCAM attribution map highlighting image regions supporting the bounding-box prediction. (C) HiResCAM attribution with spatial alignment to the object. (D) LIME shows superpixels with the highest estimated contribution to the prediction.

Figure 3 shows a true positive and false positive harbor seal detection using the Faster-RCNN model. The top detection is a true positive seal, whereas the bottom detection is a dark ice patch incorrectly classified as a seal (false positive). The gradient-based explanations shown in Figure 3B (LayerCAM) and Figure 3C (HiResCAM) highlight regions of the dark ice patch that the model attributes to being predicted as a seal. Specifically, by comparing the regions highlighted in our true positive and false positive, we observe that the curvature of the dark ice patch is very similar to the body of a seal. The LIME superpixels containing the seal and dark ice patch also have the highest estimated contribution to the prediction (Figure 3D). In this scenario, our explanations highlight that the model is behaving as expected, and the dark ice patch observed in this image is very similar to the curvature of a seal. Data augmentation via perturbations or additional data collection of dark ice patches could be used to improve the model and reduce false positives in this scenario. This explanation provides evidence to support the decision to retrain our model on augmented or new data. In the case of incorrect predictions, explainability techniques can help us better understand why a model made a prediction, debug our models, and help inform decisions for next steps, including data augmentation and/or additional data collection.

Perturbation-based explanations are helpful to evaluate model robustness by crafting “what if” scenarios. We can do this using diffusion model-based inpainting, where we segment out the animal and replace it with a generated background (Aghakishiyeva et al., 2025). Unlike traditional perturbations, which blur or black out image regions, inpainting perturbations keep edits in-distribution and preserve ecological semantics, so both models and experts view plausible scenes. This makes explanations more actionable: removals test whether species morphology is required for an accurate detection and replacements probe “what else looks like the target?”. Because scenes remain coherent, expert reviewers can better judge plausibility.

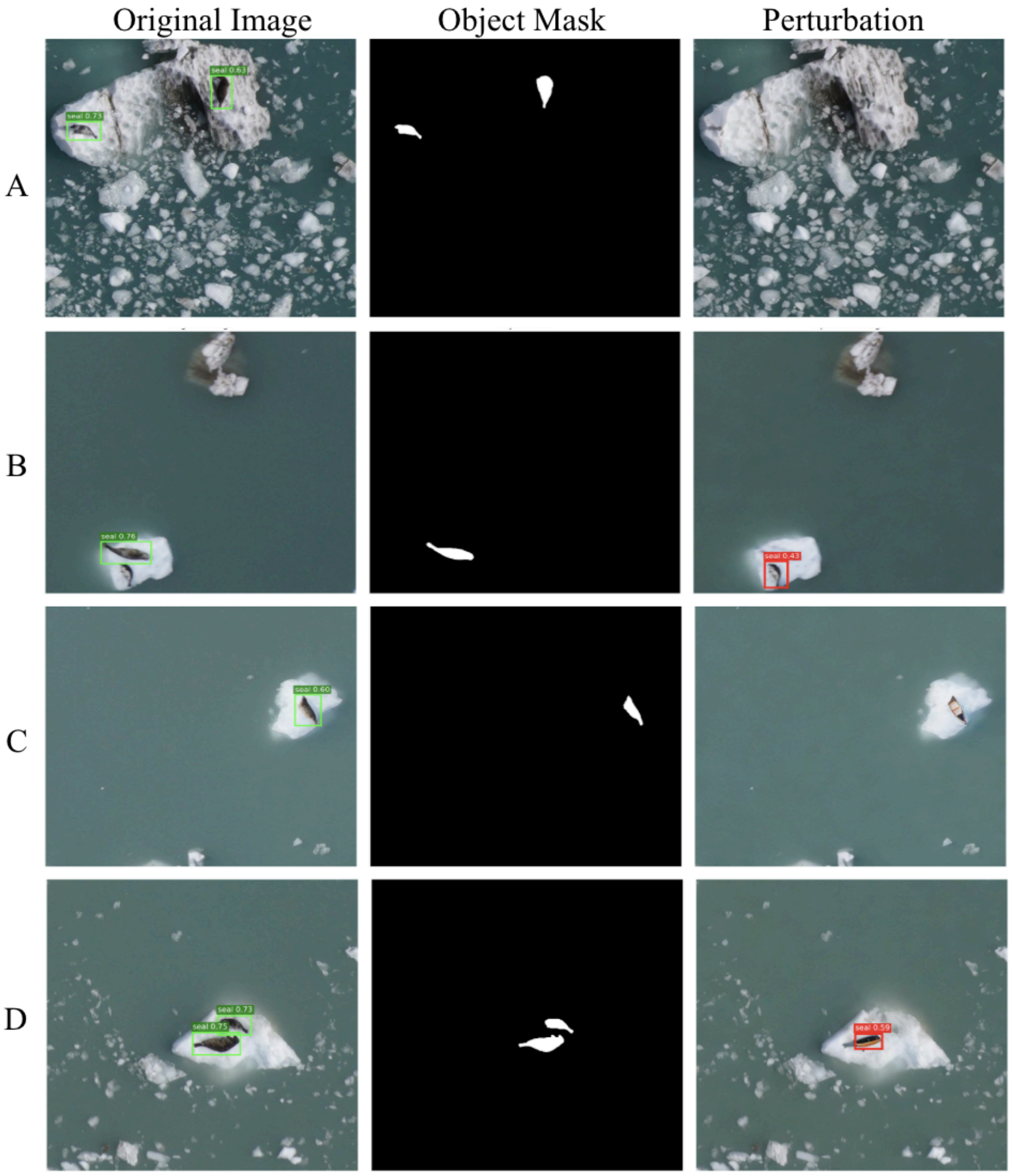

**Figure 4.** Inpainting-based perturbations of harbour seals in drone imagery and detections from YOLOv9 model. (A) Removing seals and inpainting with ice eliminates the detections. (B) Removing one seal flips its prediction while increasing confidence for a neighboring seal. (C) Replacing a seal with a boat flips the prediction. (D) The boat replacement is misclassified as a seal.

A perturbation example is shown in Figure 4, where segmented seals are replaced with background ice (A and B) or a boat (C and D). When we replace a seal with ice and the individual is no longer detected, this provides evidence that the model is utilizing the fine-grained morphology of the seal, rather than the broad context, to predict the seal (Figure 4A). We also highlight a case where removing one seal flips the prediction of the removed seal but the confidence of the detection for a neighboring seal increases (Figure 4B). This may indicate interdependence of predictions when multiple seals are present and/or limitations of our dataset. For example, having limited images with multiple seals on the same ice floe. We show an example where the replacement with a boat flipped the prediction (Figure 4C) and an example where the model still detected the boat as a seal (Figure 4D). Replacement edits can reveal important confounds. Boats that remain misclassified as seals indicate that the model is relying on overlapping edge and texture cues shared by seals and boats. Before deployment, removal edits can be used to test whether erasing key morphological characteristics suppresses detections, thereby confirming which features drive the model's predictions. Replacement edits can then highlight hard negatives, such as boats, that should be prioritized for targeted data curation and augmentation to improve model discrimination.

### 5.2 Case Study: Cetacean Anatomical Region Segmentation

To illustrate an explainability technique with a segmentation use case, we trained a YOLOv8 Model (Yeh et al., 2024) (Appendix A.3) on a dataset of drone images of three cetacean species: minke whales, blue whales, and humpback whales (Gray et al., 2019). The segmentation model was trained to identify and outline whale bodies and pectoral fins. To investigate model uncertainty, we used LIMEs adapted for use with segmentation models (Appendix A.4.2).

Figure 5A depicts the segmentation of two individual whales. LIME attributions concentrate on the whales themselves, with secondary importance on the water disturbance around the upper whale, indicating that the model relies on both morphology and local water texture. This is consistent with expectations but also flags a potential dependence on contextual cues. Figure 5B illustrates a failure case which clarifies model limits, with a missed submerged whale receiving weak attributions, consistent with low contrast and partial occlusion. In Figure 5C, an object at the frame edge elicits spurious importance and a false positive, suggesting potential boundary effects. In a test image with a boat (Figure 5D), the model correctly segments only the whale, and LIME assigns negligible importance to the boat region, indicating some robustness to anthropogenic distractors.

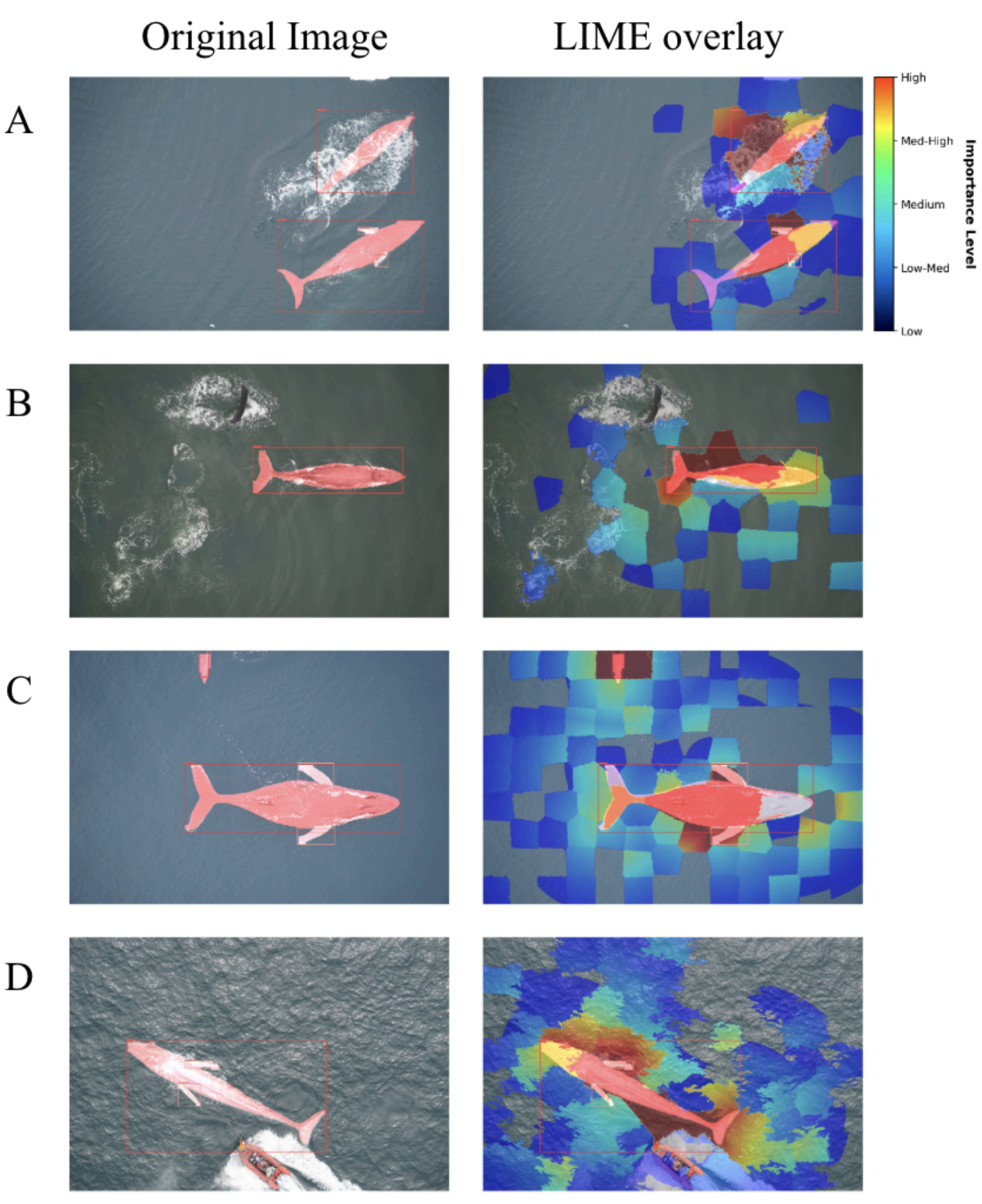

**Figure 5.** LIME shows superpixels with the highest estimated contribution to the prediction. (A) Segmentation of multiple whales shows importance of water patterns, (B) Missed submerged whale, (C) False positive, (D) Expected model behavior when a boat is present.

Explainability techniques for segmentation models, like LIME, enable a check of whether attributions align with biologically meaningful structures (e.g., animal morphology) versus background textures, which could indicate model fragility. Explanations also inform actionable next steps around data augmentation and curation, by highlighting non-target regions (e.g., boat wakes) and near misses (submerged individuals, edge crops) that can be targeted in the next iteration of model development.

## 6. Implementation in Ecological Workflows

While there is not yet a comprehensive software platform for explanations in ecology, open-source software for the implementation of many explainability approaches does exist, and we recommend the use of these for initial implementations. These tools may require some modification for certain workflows, especially for image segmentation tasks or for use in state of the art models. The *pytorch-grad-cam* library (Gildenblat, 2021) provides excellent documentation and clear recommendations for applying CAM methods to popular computer vision models, including Faster R-CNN, ResNet, VGG, Vision Transformer (ViT), and Swin Transformer (SwinT) and the *lime* python library (Ribeiro, 2020) offers comprehensive documentation and well-designed functions to generate LIME explanations.

## 7. Future Directions

Responsible use of AI in ecological research requires more than accurate predictions. Whenever possible, researchers should consider interpretable-by-design models (Rudin, 2019). When complex but opaque models are necessary, post-hoc explainability should be integrated into evaluation workflows alongside conventional assessments of uncertainty, robustness, calibration, external validation, and expert ecological review.

### 7.1 Ecological XAI Benchmarks

A major barrier to progress is the lack of ecological benchmarks for evaluating explanation methods. Future work should develop benchmark datasets that include common ecological computer vision tasks along with known sources of variation. These benchmarks should allow researchers to compare explanation methods using shared models, datasets, and evaluation protocols.

### 7.2 Expert-Grounded Evaluation

Evaluating XAI remains difficult because there is often no ground-truth explanation for a model prediction, highlighting the need for input from domain experts (Wickstrom et al., 2024). Future studies should include expert-grounded evaluation, in which ecologists assess whether explanations align with biologically meaningful traits, ecological context, and known sources of uncertainty. These evaluations should test whether explanations help experts identify model errors, detect dataset bias, improve annotation protocols, select additional training data, or make better deployment decisions. Human evaluation of XAI remains underdeveloped across the broader field, making ecology an important domain for advancing application-grounded assessment (Nauta et al., 2023; Suh et al., 2025).

### 7.3 Transformer and Foundation-Model Explainability

Transformer-based and foundation models are increasingly used in ecological applications, offering new capabilities but creating new explainability challenges. Future work should evaluate whether attention-based explanations, prompt sensitivity analyses, and mask-stability tests provide reliable information in ecological imagery. For segmentation foundation models, researchers should assess whether outputs remain stable across plausible point, box, and mask prompts, and whether unstable masks correspond to ecologically ambiguous boundaries or model failure.

### 7.4 Spatial-Temporal and Multimodal Ecological Data

Most current XAI workflows explain single images, yet ecological inference often depends on spatial and temporal structure. Future methods should account for repeated sampling across sites, seasonal variation, animal movement, species interactions, habitat change, and sensor differences. Explanations should also be extended to multimodal ecological systems that combine imagery with audio, telemetry, environmental covariates, weather, remote sensing products, or text-based observations. These settings require methods that can explain not only which image regions influenced a prediction, but also how information from multiple data streams contributed to a prediction.

### 7.5 Uncertainty, Calibration, and Robustness

Explainability should not be treated as a substitute for conventional uncertainty estimation or robustness testing. A model can produce a convincing explanation while still being poorly calibrated, sensitive to domain shift, or unreliable under novel environmental conditions. Future ecological XAI workflows should combine explanations with uncertainty estimates, calibration checks, out-of-distribution testing, and external validation across sites or seasons. This is particularly important for applications where false positives or false negatives may have substantial ecological or social consequences.

### 7.6 Open-Source Ecological XAI Workflows

Broader adoption by researchers will require accessible open-source tools designed around ecological data and workflows. Existing XAI libraries provide useful starting points, but they often require adaptation for ecological tasks. Future tools should make it easier to generate prediction-level and dataset-level explanations, aggregate attribution across ecological covariates, compare explanations with masks or bounding boxes, run perturbation tests, and export results for expert review. Cross-disciplinary collaboration among ecologists, computer vision researchers, human-computer interaction researchers, and conservation practitioners will be essential for building tools that are technically sound and practically useful.

Together, these directions point toward a more rigorous role for explainability in ecological AI. XAI should become part of an iterative ecological modeling workflow: diagnosing model behavior, identifying confounds, guiding data collection, supporting expert review, and determining whether AI systems are appropriate for deployment in real ecological settings.

## 8. Conclusion

As AI systems increasingly inform conservation decisions, monitor vulnerable species, and guide resource allocation, explainability is becoming an important component of responsible ecological modeling. Ecological systems are complex, variable, and context-dependent, so model evaluation must go beyond aggregate performance metrics. Researchers need tools for examining both individual predictions and broader dataset-level patterns of model behavior. Without this scrutiny, models may appear accurate while relying on biased, unstable, or ecologically implausible cues. By integrating explainability, ecologists can better interrogate model behavior, identify errors and confounds, and evaluate whether model outputs align with domain expertise. XAI provides practical tools for improving model development, strengthening ecological inference, and supporting more responsible management of the natural environment.

## References

**Note on references:** In computer science and machine learning, many high-impact and foundational contributions are published in double-blind peer-reviewed conference and workshop proceedings rather than traditional journals. Accordingly, several key references cited in this work appear in conference proceedings that serve as primary archival venues within these fields.